\PassOptionsToPackage{unicode}{hyperref}
\PassOptionsToPackage{hyphens}{url}
\PassOptionsToPackage{dvipsnames,svgnames,x11names}{xcolor}
\documentclass[
]{article}

\usepackage{amsmath,amssymb}
\usepackage{iftex}
\ifPDFTeX
  \usepackage[T1]{fontenc}
  \usepackage[utf8]{inputenc}
  \usepackage{textcomp} % provide euro and other symbols
\else % if luatex or xetex
  \usepackage{unicode-math}
  \defaultfontfeatures{Scale=MatchLowercase}
  \defaultfontfeatures[\rmfamily]{Ligatures=TeX,Scale=1}
\fi
\usepackage{lmodern}
\ifPDFTeX\else  
\fi
\IfFileExists{upquote.sty}{\usepackage{upquote}}{}
\IfFileExists{microtype.sty}{% use microtype if available
  \usepackage[]{microtype}
  \UseMicrotypeSet[protrusion]{basicmath} % disable protrusion for tt fonts
}{}
\makeatletter
\@ifundefined{KOMAClassName}{% if non-KOMA class
  \IfFileExists{parskip.sty}{%
    \usepackage{parskip}
  }{% else
    \setlength{\parindent}{0pt}
    \setlength{\parskip}{6pt plus 2pt minus 1pt}}
}{% if KOMA class
  \KOMAoptions{parskip=half}}
\makeatother
\usepackage{xcolor}
\ifx\paragraph\undefined\else
  \let\oldparagraph\paragraph
  \renewcommand{\paragraph}[1]{\oldparagraph{#1}\mbox{}}
\fi
\ifx\subparagraph\undefined\else
  \let\oldsubparagraph\subparagraph
  \renewcommand{\subparagraph}[1]{\oldsubparagraph{#1}\mbox{}}
\fi

\usepackage{longtable,booktabs,array}
\usepackage{calc} % for calculating minipage widths
\usepackage{etoolbox}
\makeatletter
\patchcmd\longtable{\par}{\if@noskipsec\mbox{}\fi\par}{}{}
\makeatother
\IfFileExists{footnotehyper.sty}{\usepackage{footnotehyper}}{\usepackage{footnote}}
\makesavenoteenv{longtable}
\usepackage{graphicx}
\makeatletter
\def\maxwidth{\ifdim\Gin@nat@width>\linewidth\linewidth\else\Gin@nat@width\fi}
\def\maxheight{\ifdim\Gin@nat@height>\textheight\textheight\else\Gin@nat@height\fi}
\makeatother
\setkeys{Gin}{width=\maxwidth,height=\maxheight,keepaspectratio}
\makeatletter
\def\fps@figure{htbp}
\makeatother
\newlength{\cslhangindent}
\newlength{\csllabelwidth}
\newlength{\cslentryspacingunit} % times entry-spacing
\newenvironment{CSLReferences}[2] % #1 hanging-ident, #2 entry spacing
 {% don't indent paragraphs
  \setlength{\parindent}{0pt}
  \ifodd #1
  \let\oldpar\par
  \def\par{\hangindent=\cslhangindent\oldpar}
  \fi
  \setlength{\parskip}{#2\cslentryspacingunit}
 }%
 {}
\usepackage{calc}

\usepackage{arxiv}
\usepackage{orcidlink}
\usepackage{amsmath}
\usepackage[T1]{fontenc}
\makeatletter
\@ifpackageloaded{caption}{}{\usepackage{caption}}
\AtBeginDocument{%
\ifdefined\contentsname
  \renewcommand*\contentsname{Table of contents}
\else
  \newcommand\contentsname{Table of contents}
\fi
\ifdefined\listfigurename
  \renewcommand*\listfigurename{List of Figures}
\else
  \newcommand\listfigurename{List of Figures}
\fi
\ifdefined\listtablename
  \renewcommand*\listtablename{List of Tables}
\else
  \newcommand\listtablename{List of Tables}
\fi
\ifdefined\figurename
  \renewcommand*\figurename{Figure}
\else
  \newcommand\figurename{Figure}
\fi
\ifdefined\tablename
  \renewcommand*\tablename{Table}
\else
  \newcommand\tablename{Table}
\fi
}
\@ifpackageloaded{float}{}{\usepackage{float}}
\floatstyle{ruled}
\@ifundefined{c@chapter}{\newfloat{codelisting}{h}{lop}}{\newfloat{codelisting}{h}{lop}[chapter]}
\floatname{codelisting}{Listing}

\makeatother
\makeatletter
\@ifpackageloaded{caption}{}{\usepackage{caption}}
\@ifpackageloaded{subcaption}{}{\usepackage{subcaption}}
\makeatother
\makeatletter
\@ifpackageloaded{tcolorbox}{}{\usepackage[skins,breakable]{tcolorbox}}
\makeatother
\makeatletter
\@ifundefined{shadecolor}{\definecolor{shadecolor}{rgb}{.97, .97, .97}}
\makeatother
\ifLuaTeX
  \usepackage{selnolig}  % disable illegal ligatures
\fi
\IfFileExists{bookmark.sty}{\usepackage{bookmark}}{\usepackage{hyperref}}
\IfFileExists{xurl.sty}{\usepackage{xurl}}{} % add URL line breaks if available
\hypersetup{
  pdftitle={Improved K-mer Based Prediction of Protein-Protein Interactions With Chaos Game Representation, Deep Learning and Reduced Representation Bias},
  pdfauthor={Ruth Veevers; Dan MacLean},
  colorlinks=true,
  linkcolor={blue},
  filecolor={Maroon},
  citecolor={Blue},
  urlcolor={Blue},
  pdfcreator={LaTeX via pandoc}}

\newcommand{\runninghead}{A Preprint }
\renewcommand{\runninghead}{GRAND, CGRs and Deep Learners }
\title{Improved \(K\)-mer Based Prediction of Protein-Protein
Interactions With Chaos Game Representation, Deep Learning and Reduced
Representation Bias}
\def\asep{\\\\\\ } % default: all authors on same column
\author{\textbf{Ruth Veevers}~\orcidlink{0000-0001-7599-5725}\\\\The
Sainsbury Laboratory\\Norwich,\ NR4 7UH\\\asep\textbf{Dan
MacLean}~\orcidlink{0000-0003-1032-0887}\\\\The Sainsbury
Laboratory\\Norwich,\ NR4
7UH\\\href{mailto:dan.maclean@tsl.ac.uk}{dan.maclean@tsl.ac.uk}}
\date{}
\begin{document}
\maketitle
\begin{abstract}
Protein-protein interactions drive many biological processes, including
the detection of phytopathogens by plants' R-Proteins and cell surface
receptors. Many machine learning studies have attempted to predict
protein-protein interactions but performance is highly dependent on
training data; models have been shown to accurately predict interactions
when the proteins involved are included in the training data, but
achieve consistently poorer results when applied to previously unseen
proteins. In addition, models that are trained using proteins that take
part in multiple interactions can suffer from representation bias, where
predictions are driven not by learned biological features but by
learning of the structure of the interaction dataset.

We present a method for extracting unique pairs from an interaction
dataset, generating non-redundant paired data for unbiased machine
learning. After applying the method to datasets containing
\emph{Arabidopsis thaliana} and pathogen effector interations, we
developed a convolutional neural network model capable of learning and
predicting interactions from Chaos Game Representations of proteins'
coding genes.
\end{abstract}
\ifdefined\Shaded\renewenvironment{Shaded}{\begin{tcolorbox}[breakable, enhanced, borderline west={3pt}{0pt}{shadecolor}, sharp corners, frame hidden, interior hidden, boxrule=0pt]}{\end{tcolorbox}}\fi

\hypertarget{introduction}{%
\section{Introduction}\label{introduction}}

Phytopathogens are a major threat to global crop production. The fungal
phytopathogen \emph{Magnoporthe oryzae} that causes cereal blast is
responsible for around 30\% of rice production loss and has now emerged
as a pandemic problem on wheat (Nalley et al. 2016). The oomycete
\emph{Phytophthora infestans} causes losses of around 6 billion USD to
potato production, annually (Haas et al. 2009). The bacterium
\emph{Ralstonia solanacearum} has a wide host range and can cause loses
of over 30\% in potato, banana and groundnut (Yuliar, Nion, and Toyota
2015). The incidences of crop disease are increasing as global climate
change and agricultural practice are expanding the geographical range of
pathogens and upping the stakes in the evolutionary arms race.

Plant detection of pathogen infection occurs at two levels: the cell
surface where pathogen molecules (PAMPS) are detected by a family of
Receptor Like Kinase (RLK) cell surface receptors to trigger immune
response and also intracellularly where pathogen effectors can interact
with R-Proteins that trigger the immune response. Identifying
interacting RLK/PAMPs and Effector/R-Proteins is a key part of immunity
research. On the plant side, RLKs and R proteins are well characterised
and can be identified bioinformatically. On the pathogen side PAMPs and
effector proteins are the shock troops of infection, manipulating the
host at the infection interface to the pathogens advantage. Identifying
and characterising a pathogen's effector content is a critical first
step in understanding diseases and developing resistance, but until
recently effectors were notoriously difficult to characterise from
sequence data. In most phyla they have only a few easily determined
sequence characteristics (some in fungi are cysteine rich or have a MAX
motif, some in oomycetes have the RXLR motif or WY fold) but in many
cases no sequence identifiers are known (Franceschetti et al. 2017).

To understand infection processes, to provide genome-level understanding
of the functions of this important class of genes and to develop future
disease resisting crop varieties and agricultural management strategies
there is a critical need to identify PAMP-RLK and effector-R gene
complements and immune triggering pairings computationally from genome
and protein sequence data.

Kristianingsih and MacLean (2021) recently developed class-leading deep
learning based models for effectors that perform classification of
effectors across species and phyla with an accuracy greater than 90\%.
This removed the need for older computational pipelines with in-built
\emph{a priori} assumptions about what might constitute an effector
sequence in the absence of sequence features known to group them
(Sperschneider et al. 2015). A significant aspect of these models is
that training set sizes were on the order of 100-200 input sequences,
demonstrating that with appropriately parameterized and optimized
architectures, deep learning can be successfully applied despite
apparently low sample sizes.

\hypertarget{prediction-of-ppis}{%
\subsection{Prediction of PPIs}\label{prediction-of-ppis}}

The utility of PPIs such as these has led to the development of several
tools for the prediction of new ones in order to discover proteins that
may interact with a given target protein. Molecular dynamics simulations
can model the physics underlying proteins' behaviours to simulate how
they may interact. However, these simulations come at a high
computational cost, particularly when attempting to model an event as
specific and potentially rare as these interactions, making the method
unsuitable for large-scale searches.

There is considerably more primary sequence data than structural data
known regarding proteins, and so methods for prediction of interactions
that require only sequence data have been developed. Following the
success of AlphaFold2(Jumper et al. 2021) in predicting protein
structure from sequence, there have been attempts to apply the AlphaFold
model to other problems. AlphaFold-Multimer(Evans et al. 2021) predicts
complexes formed by multiple protein chains, and AlphaPulldown(Yu et al.
2022) builds this into means of searching for interactions in large
databases.

Sequence data has been encoded for machine learning as \(k\)-mer counts
or as one-hot encoding. Convolutional Neural Networks (CNNs) have
previously been applied to the problem. Sequence data has been augmented
using using PSI-BLAST to derive input matrices for each protein
containing probabilistic protein profiles or Position-Specific Scoring
Matrices(Hashemifar et al. 2018) (Wang et al. 2019). Approaches
including Hashemifar et al. (2018) use shared parameters in
twinned(Bromley et al. 1993) architectures, training a single model to
extract features from both components of the PPI. Ding and Kihara (2019)
also incorporates physiochemical properties like side-chain charge and
hydrophobicity.

Some methods of predicting from protein sequence have approached
sequences using methods designed for handling language: creating
doc2vec(Le and Mikolov 2014) encodings of a set of sequences to use as
features(X. Yang et al. 2020), or building a protein language model
through the use of transformers(Rao et al. 2021) (Rives et al. 2021).
Qiu et al. (2020) incorporates both homology and language vector
encoding for per-residue binding prediction.

S. Yang et al. (2019) apply existing PPI prediction methods to
interactions between \emph{Arabidopsis thaliana} and pathogen proteins
and find that the methods do not translate well. They supplement their
random forest model by explicitly including information about the
network of interactions.

\hypertarget{representation-bias}{%
\subsection{Representation bias}\label{representation-bias}}

The paired input of PPI data introduces a potential source of bias. Each
input sample consists of two proteins, so proteins could appear in
multiple pairs. Park and Marcotte (2012) enumerate the ways that
proteins can be combined into pairs: both proteins in a test pair might
be part of the training set (in-network), or one might be previously
unseen by the model, or they could both be unseen (out-of-network). Test
sets that consisted of in-network proteins yielded better performance
that those featuring out of network proteins. Hamp and Rost (2015)
identify similar groups and concluded that the choice of test group
dramatically affected the prediction results, while also identifying
that including a given protein in multiple pairs in the training data
lowers the accuracy on unseen test proteins compared to including it in
a single training pair. Eid et al. (2021) examine the extent to which
this has effected published machine learning studies, and demonstrate
that models can achieve 0.99 AUC scores on existing benchmark datasets
even when the protein sequences have been masked such that biological
features cannot contribute to prediction. They propose a comprehensive
framework for identifying bias in PPI datasets.

As well as representation bias driven learning, the application of
machine learning to biological sequences also requires consideration of
the issue of data leakage from homology(Jones 2019). Pairs of sequences
may correspond to related structures with very similar forms and
functions despite differences in sequence(Rost 1999). The inclusion of
homologous biomolecules appearing in both the testing and training data
can result in overestimation of a model's accuracy and poor
generalization to other data. While sequence identity cannot fully
identify all homologous proteins, a common approach used by recent
studies in the prediction of PPIs(X. Yang et al. 2020), binding
residues(Qiu et al. 2020), tertiary structures(Singh et al. 2019), and
other biomolecular properties is to filter by sequence identity using
software like CD-Hit(Li and Godzik 2006).

However, machine learning models are widely understood to require
abundant training data. The naive method of sequence filtering for PPI
prediction - clustering to a given sequence identity threshold and then
discarding interactions until there are no pairs of interactions that
use a protein from the same cluster - potentially discards more
information than necessary to a degree that depends on the random seed.

\hypertarget{chaos-game-representation}{%
\subsection{Chaos Game Representation}\label{chaos-game-representation}}

\begin{figure}

{\centering \includegraphics{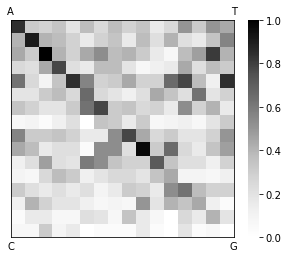}

}

\caption{\label{fig-cgr-example}An example CGR attractor, built from the
sequence of the Arabidopsis \emph{CUL4} gene (TAIR ID: AT5G46210.1) at a
4-mer resolution. This CGR incorporates frequency; the darkness of each
pixel corresponds to a given 4-mer's frequency in the sequence; each
cell contains the number of times the 4-mer appears, normalised within
each CGR image.}

\end{figure}

The Chaos Game Representation (CGR)(Jeffrey 1990) of a nucleotide
sequence is a representation of the composition of a sequence as a
2-dimensional image. The ``Chaos Game'' is a process in which a sequence
drawn from an alphabet of \(n\) characters can be represented as points
in an \(n\)-sided polygon. Certain values of \(n\) produce fractal
patterns given enough points, but when \(n=4\), the square produced is
filled uniformly. To encode a genetic sequence, the four points of a CGR
represent the four nucleotides and a point is added for each residue in
the sequence. If the square is divided into quadrants, the number of
points in each quadrant will be the number of times the nucleotide
representing the nearest corner occurs in the sequence. Further division
of the quadrants creates sub-quadrants, containing points corresponding
to the occurrences of two-character-long subsequences (or 2-mers). This
division and subdivision can be repeated to any value of \(k\). By
setting a value for \(k\), CGRs can be represented as a 2-dimensional
data structure in which each cell contains either booleans (to show
presence or absense of a corresponding \(k\)-mer) or numbers (to show
the number of times the \(k\)-mer occurs).

An example with frequency information included at 4-mer resolution is
shown in Figure~\ref{fig-cgr-example}, which shows some visually
identifiable indications of the composition of the sequence. The dark
line along the diagonal shows an abundance of 4-mers that consist solely
of ``A'' and ``G'' residues, whereas the lighter region along the bottom
edge indicates that 4-mers of ``C''s and ``G''s in combination are more
scarce.

A variety of methods have been proposed to encode sequences of amino
acids as CGRs; Dick and Green (2020) implements four of these approaches
for a taxonomic classification task: reverse mapping to standardised
nucleotide sequences(Deschavanne and Tuffery 2008); two versions of
representation via a 20-sided polygon with a point for each potential
amino acid, with appearance of 9-mers represented either as a greyscale
intensity or black-and-white binary; and a 20-flake frequency matrix
chaos game representation(Löchel et al. 2020). The models performed
similarly modestly. Jia et al. (2019) uses chaos game representation to
build inputs for a PPI classifier, converting their protein sequences
into pseudo-nucleic sequences by replacing each amino acid with a single
representative codon following Deschavanne and Tuffery (2008) and
counting frequencies of points within the regions of the CGR attractors
to pass as input into an ensemble of random forest classifiers.

The 2-dimensional arrangement of \(k\)-mers representing a sequence is
reminiscent of single channel images, and we have extended CGRs such
that each array element contains an ordered 2-tuple of \(k\)-mer counts
from 2 distinct sequences, analogous to a two-channel image. There has
been extensive research into machine learning for the purposes of
computer vision, resulting in powerful models capable of image
processing tasks such as semantic segmentation and object recognition.
When rendered as an image, a CGR displays recognisable visual patterns,
suggesting that it would be amenable to use in deep-learning models
effective in recognising patterns in images such as those based on
Convolutional Neural networks (CNNs). In this work, we develop and train
CNN architectures for the prediction of \emph{Arabidopsis thaliana} PPIs
from two-channel CGR tensors that represent the pair of gene sequences
that code for the proteins. We extend this further using transfer
learning, focusing specifically on interactions between
\emph{Arabidopsis thaliana} proteins and effectors. We also explore the
ability of our models to identify features specific to taxonomic ranks,
testing their performance by examining two metagenomics questions:

\hypertarget{augmentations-and-synthetic-data}{%
\subsection{Augmentations and Synthetic
data}\label{augmentations-and-synthetic-data}}

While the image-like format of the CGR attractors proved well-suited to
CNN models, the field of image processing using neural networks has
developed methods beyond architectural considerations to improve models'
performance on tasks. On-the-fly augmentations, from simple
transformations such as rotation and scaling to more complex blending of
images(Xu et al. 2023), have assisted in the training of object
detection and semantic segmentation neural networks. By showing the
adjusting the training data during supervised learning, the model is
exposed to more variety in its training data with the intent that it is
more readily able to generalise to unseen test data.

Additionally, various studies have demonstrated methods of generating
synthetic images. Supervised machine learning for image processing tasks
like object detection require ground truth data from which to learn, but
the per-pixel labelling of the contents of an image can be an expensive
or time-consuming bottleneck, particularly if the labelling task calls
for expert knowledge. Synthetic images present an opportunity to quickly
produce both an image and its ground truth labels and gives researchers
control over the contents of their training data which has been used to
augment real training data in some domains(Man and Chahl 2022).

Cut-and-paste methods(Dwibedi, Misra, and Hebert 2017) take small
numbers of manually labelled image components and position them on a
background to vary the composition of a collection of images. While
computationally simple, this has recently been successfully applied to
plant phenotyping (Albert et al. 2021). Even more control over an
image's contents can be gained when the image is rendered using
3-dimensional modelling, though the process of creating and rendering
the objects can still require significant time and expense. Researchers
have used video game engines Unreal Engine(Epic Games 2023) and Grand
Theft Auto V(Rockstar Games 2013) as well as bespoke engines to render
environments(Saleh et al. 2018). Ward and Moghadam (2020) create 3D
plants to render as top-down images along with segmentation masks for
each leaf. Generative Adversarial Nets (GANs)(Goodfellow et al. 2014)
create images using a pair of specialised and opposed neural networks,
where one attempts to generate images that can pass as real and another
attempts to identify whether an image is real. Each model improves as it
trains which makes the other model's task harder, driving improvement
further. Shin et al. (2018) leverage this process not just to generate
more training data but to anonymise medical photographs. The GAN model
can be extended to Conditional Generative Adversarial Nets (cGANS)(Mirza
and Osindero 2014) to generate images belonging to different classes,
such as photographs of diseased and non-diseased plants(Abbas et al.
2021).

Training data for PPI models must be experimentally determined, which
requires time, expense and expert knowledge. As such, exploiting
existing data to further training poses a potential advantage. However,
despite their image-like properties, standard methods of image
augmentation such as rotation, cropping and scaling would not work for
CGR representations as they would change the underlying sequences such
that the labels could no longer reliably be used as ground truth. We
explore two methods of augmenting or generating synthetic CGR data.

\hypertarget{results}{%
\section{Results}\label{results}}

\hypertarget{graph-reduction-algorithm-for-non-redundant-datasets-grand-consistently-retains-more-data-than-random-discarding-of-duplicated-samples}{%
\subsection{Graph-Reduction Algorithm for Non-redundant Datasets (GRAND)
consistently retains more data than random discarding of duplicated
samples}\label{graph-reduction-algorithm-for-non-redundant-datasets-grand-consistently-retains-more-data-than-random-discarding-of-duplicated-samples}}

The issue of redundancy in biological sequence data caused by high
sequence similarity is exacerbated in the case of PPI predictions, where
each sample is composed of two sequences. The representation bias-driven
learning that this can introduce may result in models that appear to
perform well but which cannot generalise to new data. We enforce strict
non-redundancy in the training and test datasets to remove this
potential source of bias, to seek a model that can predict interactions
from unseen, out-of-network sequences. However, a secondary priority is
that since labelled, experimentally-derived interaction data is not
trivial to obtain, we wish to retain as many samples as possible.

\begin{figure}

{\centering \includegraphics{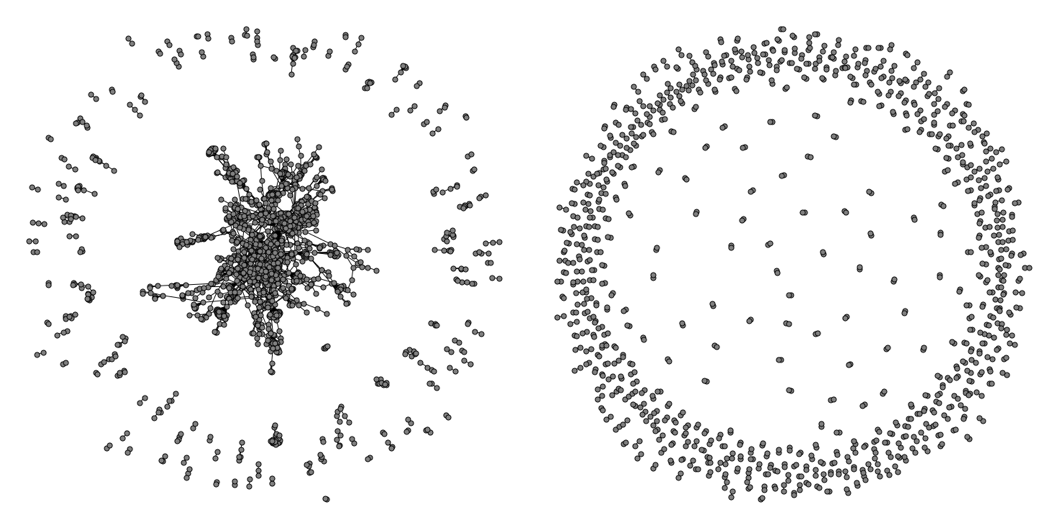}

}

\caption{\label{fig-grand-effect}(a) The network of Arabidopsis thaliana
PPIs from which the positive samples were selected. Nodes are clusters
of similar genes, and edges represent a known interaction between one or
more proteins. (b) The network after filtering with GRAND.}

\end{figure}

Figure~\ref{fig-grand-effect} (a) shows the PPIs from ArabidopsisPPI
represented as a network. Each node represents a cluster of one or more
similar genes, from the output of CD-Hit. Each edge represents a known
interaction between at least one pair of the proteins coded for by the
genes contained within each cluster. A node's degree is defined as the
number of other nodes to which it is connected by an edge. Many of the
genes' products are included in multiple interactions; some nodes in the
network have a degree of over one hundred. GRAND iteratively prunes this
network until each remaining node has one single neighbour.

While any node has a degree greater than one, the following is repeated.
Each node with degree one is taken in turn, and its sole neighbour is
found. If the neighbour has other edges, they are removed. When no nodes
with degree one remain, for each edge an ``edge-sum'' is calculated by
summing together the degrees of the two nodes that it connects. The edge
with the lowest edge-sum is selected for inclusion, and all other edges
between the connected nodes and their other neighbours are dropped. Any
nodes that are no longer connected by any edges are removed, and
considered for inclusion as part of a negative pair. The dataset is then
constructed from this network by choosing randomly one of the known
interactions that each edge represents.

Figure~\ref{fig-grand-effect} (b) shows the new network, wherein each
node shares an edge with exactly one other node. This new network
contains 1,106, whereas the network created by random discarding
contained 867. This is a 27.6\% increase in available non-redundant
data.

The improvement holds for the other datasets to which we have applied
GRAND. On the EffectorK host/pathogen effector interaction dataset,
which contains 1,220 interactions between 769 CD-Hit clusters, 1,000
applications of the naive approach returns a mean of 131.62 interactions
(standard deviation 5.26), where none of the attempts return more than
148 interactions. GRAND returns 183 interacting pairs, for a 23.6\%
increase over the maximum naive result and 39.0\% increase over the
mean. The larger HPIDB dataset contains 36,139 interactions between
13,099 CD-Hit clusters. 100 applications of the naive approach returned
between 1,796 and 2,010 pairs, with a mean of 1,893.23 and a standard
deviation of 34.29. GRAND finds 3,282 non-redundant pairs, which
increases the non-redundant data by 63.3\% over the maximum naive result
and 73.4\% over the mean.

GRAND offers a fast, standard process for creating non-redundant
datasets that consistently contain more data than is achieved by a
random approach, offering increases of between 23.6\% and 73.4\% more
paired samples in the datasets to which we applied it.

\hypertarget{cnn-models-can-learn-ppi-prediction-from-non-redundant-datasets}{%
\subsection{CNN models can learn PPI prediction from non-redundant
datasets}\label{cnn-models-can-learn-ppi-prediction-from-non-redundant-datasets}}

Knowing that out-of-network PPI prediction is a difficult task, we
anticipated that finding the right model architecture to learn from our
ArabidopsisPPI dataset would be important. We conducted a wide random
search of the model hyperparameter space in which we varied which layers
we used, and also parameters controlling the tightness of fit including
dropout rate, regularisation parameters and CGR resolution.

We selected the structure for our CNN architecture using a random grid
search of a wide parameter space. The structures followed the same
pattern: the channels were split and passed through a convolutional
stage consisting of zero or more ``convolutional blocks'' which each
contained one or more convolutional layers. As in Hashemifar et al.
(2018), we used a single model for this convolutional stage allowing a
single set of model weights to be trained on both interacting partners'
sequences. As our CGR attractors are ``image-like'', we examined
well-known and well-performing models from the field of image processing
as they were implemented in the Keras Python package, such as ResNet(He
et al. 2015), EfficientNet(Tan and Le 2019) and VGG19(Simonyan and
Zisserman 2014). Such models proved unsuitable for immediate use as the
dimensions of our CGRs were too small for the models' depth and
convolutional size, but we took inspiration from the ResNet(He et al.
2015) architecture by incorporating potential ``skip'' layers within
these convolutional blocks when they contained more than one layer; this
structure has performed well for computer vision tasks, addressing the
``vanishing gradient'' issue. Following this convolutional stage the
outputs for each input channel are pooled and concatenated to create a
single vector for each sample. A dense stage follows, wherein the vector
is passed through zero or more dense layers. A final classification
layer with sigmoid activation provides the model's output, and binary
cross-entropy is used as the model's loss function. We searched the
space by varying the numbers and depths of these layers, convolutional
filter sizes, dropout rates and regularisation parameters, as well as
the choice of \(k\) which dictates the size and resolution of the
initial input. For each value of \(k\), the same 1,000 models were
tested 5 times by training on 60\% of the dataset, using a 20\% split
for validation purposes. The same 20\% was withheld each time for final
testing after the models trained and the final models were selected
based on validation performance. Each model was trained for a maximum of
500 epochs, with early stopping triggered if the validation loss stopped
improving, and a model checkpoint kept track of the version that
achieved the best validation results.

\begin{figure}

{\centering \includegraphics{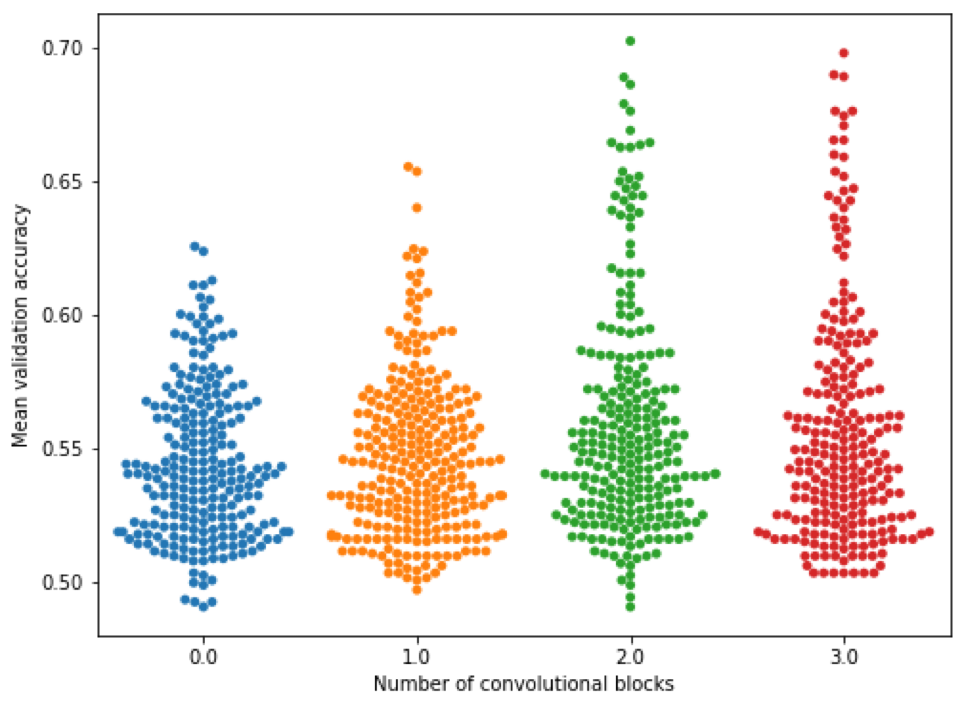}

}

\caption{\label{fig-val-acc-vs-conv}Mean validation accuracy from a
ResNet random grid architecture search using a dataset of 4-mers, with
increasing convolutional blocks}

\end{figure}

The validation accuracy varied from around 50\%, at which point the
model is indistinguishable from random, up to 70.2\%. The amount of
models with poor validation performance may indicate the difficulty of
generalisable learning in this task. We examined the relationships
between various parameters and their results, such as the number of
convolutional blocks as shown in Figure~\ref{fig-val-acc-vs-conv}. We
found that no models that skipped the convolutional stage were able to
achieve above 65\%, while the best results were obtained by models that
included two or three convolutional blocks.

\begin{figure}

{\centering \includegraphics{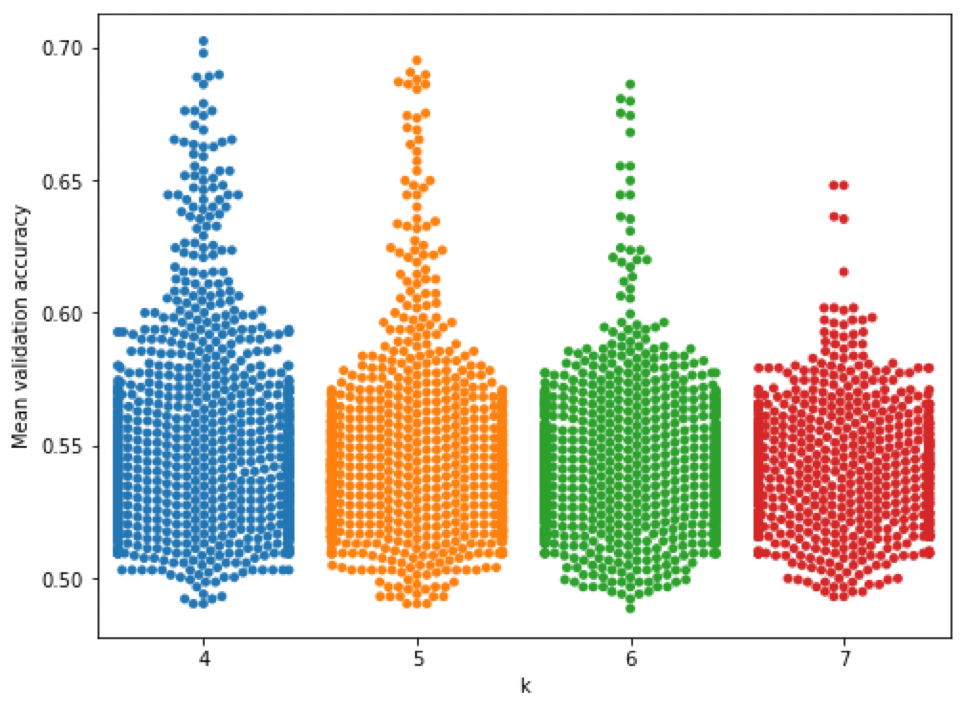}

}

\caption{\label{fig-k-vs-acc-grid-resnet}Mean validation accuracy from a
random grid search of ResNet architectures across a range of k lengths}

\end{figure}

We also studied the effect of \(k\) on accuracy in our ResNet-inspired
models. The results, as shown in Figure~\ref{fig-k-vs-acc-grid-resnet},
show that the smaller, lower-resolution 4-mers yield better validation
accuracy than higher values of \(k\).

The top 25 models from the 4,000 models considered in the architecture
search were carried forward to testing with the remaining 20\% of the
data. While these models achieved mean validation accuracies of up to
70.2\% (see Figure~\ref{fig-k-vs-acc-grid-resnet}), the highest mean
accuracy achieved on the holdout data was 65.0\%; through the use of an
ensemble of different models compiled with a meta-learner, this was
increased to 66.7\%.

\hypertarget{alternatives-to-deep-learning-do-not-achieve-comparable-accuracy}{%
\subsection{Alternatives to deep learning do not achieve comparable
accuracy}\label{alternatives-to-deep-learning-do-not-achieve-comparable-accuracy}}

To explore whether our deep learning approach was necessary, we applied
two non-Neural Network machine learning methods, support vector
classifiers and random forest classifiers, to the same ArabidopsisPPI
dataset formatted as vectors of \(k\)-mer counts. In each case, we
performed a grid search using training and validation splits to find the
hyperparameters that achieved the highest validation accuracy, then
applied those trained models to the 20\% of samples that were held out
for testing. Again, larger values of \(k\) yielded worse results;
validation accuracies achieved using 5-mer, 6-mer and 7-mer counts
appeared indistinguishable from random chance, ranging from 42.5\% to
54.3\%. The best results were achieved by the support vector classifier
using a radial basis function kernal on vectors of 3-mers, which reached
a validation accuracy of 58.4\%. However, this failed to generalise to
the hold-out data as the test accuracy was 52.3\%.

\hypertarget{transfer-learning-allowed-modest-learning-from-a-related-but-very-small-dataset}{%
\subsection{Transfer learning allowed modest learning from a related but
very small
dataset}\label{transfer-learning-allowed-modest-learning-from-a-related-but-very-small-dataset}}

We applied our methods to our EffectorPPI dataset in order to focus
specifically on the interactions between R-Proteins and
\emph{Arabidopsis thaliana} proteins.

To begin, we repeated the hyperparameter search repeated using training
and validation subsets taken from our EffectorPPI dataset. Some
parameter sets appeared to perform well, with one model architecture
achieving a mean validation accuracy of 73.9\%. However, when the best
of these models according to validation accuracy were applied to the
holdout test data, they failed to achieve any test accuracy higher than
55.4\%, close to the expected performance of a random classifier on the
50\% positive, 50\% negative data.

As this contains approximate one sixth of the number of samples in
ArabidopsisPPI, we explored tactics for using what the models had
previously learned from the comparatively abundant ArabidopsisPPI
dataset. The first method used the results of the ArabidopsisPPI
architecture, taking the hyperparameter sets that achieved the ten
highest validation accuracy scores when trained and validated using
ArabidopsisPPI We then initialised these models with random weights and
trained, validated and tested them on the EffectorPPI data. The best
mean holdout accuracy obtained by one of these randomly initialised
models was 0.576.

In our second approach, we used the same ten architectures found using
the ArabidopsisPPI parameter search, but instead of randomly
initialising new instances of these models, we used the five model
checkpoints that were produced as each model was repeatedly trained with
ArabidopsisPPI. These were used as the starting point for training and
validating on subsets of EffectorPPI, and the highest achieved mean
holdout accuracy was 60.0\%.

The final transfer learning approach we took used the same ten parameter
sets, and again started with the model checkpoints trained on
ArabidopsisPPI. We froze the weights in all but the final classification
layer of each model, before retraining using the same subsets of
EffectorPPI. The models trained in this manner achieved up to 0.597 test
accuracy.

Across the top ten models by validation, the second approach (training
on ArabidopsisPPI, then retraining all layers on EffectorPPI) yielded
generally higher holdout results, outperforming the models trained from
random initialisation for 6 of the 10 architectures, and outperforming
the models where only the classification layer was retrained in 7 of the
10 cases. However, the performance ranking varied between architectures,
and the single model with the most consistent good performance was
trained using the weight-freezing approach; each of the holdout tests
achieved an accuracy between 0.581 and 0.635.

\hypertarget{the-architecture-extends-to-other-data-and-achieves-state-of-the-art-results-on-difficult-benchmark-cases}{%
\subsection{The architecture extends to other data and achieves
state-of-the-art results on difficult benchmark
cases}\label{the-architecture-extends-to-other-data-and-achieves-state-of-the-art-results-on-difficult-benchmark-cases}}

\hypertarget{tbl-vs-ppi}{}
\begin{longtable}[]{@{}lrr@{}}
\caption{\label{tbl-vs-ppi}outperforms existing PPI
papers}\tabularnewline
\toprule\noalign{}
Model & Negatome & Recombine.pairs \\
\midrule\noalign{}
\endfirsthead
\toprule\noalign{}
Model & Negatome & Recombine.pairs \\
\midrule\noalign{}
\endhead
\bottomrule\noalign{}
\endlastfoot
Best published F1 score & 0.87 & 0.68 \\
Our F1 score (Ensemble) & 0.90 & 0.79 \\
\end{longtable}

We wanted to compare the models we had found with models in the existing
literature by using a benchmark PPI dataset. Wei et al. (2017) provides
a set of positive PPI pairs and three alternative sets of negative PPI
pairs using different strategies. We used the positive data along with
the Negatome negative data for one test and the positive data and the
RecombinePairs negative data for another.

We found the best ensemble of 10 high-performing models according to the
ArabidopsisPPI validation dataset and applied it to each dataset.
Table~\ref{tbl-vs-ppi} shows the F1 scores achieved by this ensemble on
each set. Distinguishing between the positive data and Negatome negative
data appears to be comparatively easy; the F1 scores published in Wei et
al. (2017) reached 0.87, and our model performed similarly at 0.90. The
RecombinePairs negative data provides a much more difficult challenge as
the positive and negative data contain very similar samples. Some of the
methods tested in Wei et al. (2017) obtained close to random
performance, while they were able to achieve an F1 score of 0.68 with
their highest-performing model. Our ensemble model achieved an F1 score
of 0.79, which has not been surpassed as far as we could find among
other works citing Wei et al. (2017). This suggests our model offers an
improvement over previous methods of learning PPI prediction,
particularly where the data is challenging to discriminate between.

\hypertarget{strategies-for-augmentations-and-synthetic-data-remain-unsolved}{%
\subsection{Strategies for augmentations and synthetic data remain
unsolved}\label{strategies-for-augmentations-and-synthetic-data-remain-unsolved}}

We have exploited the image-like arrangement of \(k\)-mers in CGR
attractors to develop model architectures inspired by research in the
computer vision field. Our ResNet-inspired twinned architecture models
were able to learn from non-redundant data and make predictions on
out-of-network PPIs with state-of-the-art accuracy. However, the
non-redundant datasets produced by GRAND are still comparably small by
the standards of deep learning where benchmark sets like ImageNet(Deng
et al. 2009) contain millions of training images. Typical computer
vision methods for augmenting training data manipulate images in each
training cycle, to increase the variety of training data seen by a
model. These methods are not suited for application to CGR attractors as
augmentation such as rotating, cropping, and recolouring CGR images
would change the \(k\)-mer counts they represent, destroying the
original biological sequence data. We explored two techniques to
generate new data from samples in the ArabidopsisPPI training data. In
the first method, synonymous substitution, we replace codons in the
coding region of a gene sequence with other codons such that the
produced protein still has the same sequence of amino acids. The second
method uses generative deep learning to create CGR-like images.

\hypertarget{synonymous-substitution-produces-cgrs-with-extremely-likely-labels-but-does-not-improve-model-performance}{%
\subsubsection{Synonymous substitution produces CGRs with extremely
likely labels, but does not improve model
performance}\label{synonymous-substitution-produces-cgrs-with-extremely-likely-labels-but-does-not-improve-model-performance}}

\begin{figure}

{\centering \includegraphics{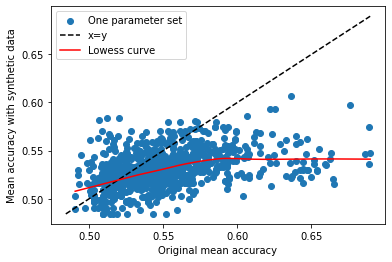}

}

\caption{\label{fig-synonymous-mutations-results}Scatter plot comparing
validation accuracy achieved by each parameter set when trained on data
with and without synthetic data generated by synonymous substitutions
included in the training data.}

\end{figure}

We generated 10 synthetic samples for each real interaction pair in our
ArabidopsisPPI 4-mers dataset using the synonymous substitution method,
and trained and validation the same parameter sets with the synthetic
training data included. As Figure~\ref{fig-synonymous-mutations-results}
shows, the highest validation accuracies were achieved without the use
of synthetic data. Fitting a Lowess curve elucidates this further: when
the model originally performed essentially randomly, the synthetic data
was in some cases able to raise the validation accuracy, but as the
original performance improves the synthetic data more often results in
worse performance. Only 23.2\% of models were more accurate with the
addition of synthetic data, and the best validation accuracy achieved
among these was only 0.61.

\hypertarget{conditional-gans-models-can-produce-cgrs-that-appear-real-but-reduce-models-accuracy}{%
\subsubsection{Conditional GANs models can produce CGRs that appear
real, but reduce models'
accuracy}\label{conditional-gans-models-can-produce-cgrs-that-appear-real-but-reduce-models-accuracy}}

\begin{figure}

\begin{minipage}[t]{0.50\linewidth}

{\centering 

\raisebox{-\height}{

\includegraphics{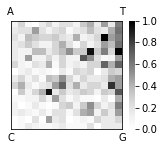}

}

}

\subcaption{\label{fig-gans-1}Real sample (channel 1)}
\end{minipage}%
\begin{minipage}[t]{0.50\linewidth}

{\centering 

\raisebox{-\height}{

\includegraphics{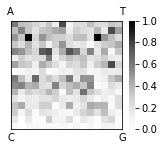}

}

}

\subcaption{\label{fig-gans-2}Real sample (channel 2)}
\end{minipage}%
\newline
\begin{minipage}[t]{0.50\linewidth}

{\centering 

\raisebox{-\height}{

\includegraphics{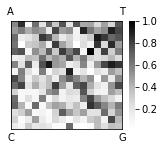}

}

}

\subcaption{\label{fig-gans-3}Synthetic sample (channel 1)}
\end{minipage}%
\begin{minipage}[t]{0.50\linewidth}

{\centering 

\raisebox{-\height}{

\includegraphics{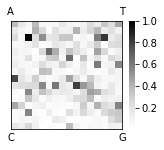}

}

}

\subcaption{\label{fig-gans-4}Synthetic sample (channel 2)}
\end{minipage}%

\caption{\label{fig-gans-combined}GAN outputs}

\end{figure}

Figure~\ref{fig-gans-combined} shows real and synthetic CGR attractors.
The real images (Figure~\ref{fig-gans-1} and Figure~\ref{fig-gans-2})
represent the two proteins interacting in a real PPI. The synthetic
images (Figure~\ref{fig-gans-3} and Figure~\ref{fig-gans-4}) were
generated using a cGAN model as examples of the true interacting class.

\begin{figure}

{\centering \includegraphics{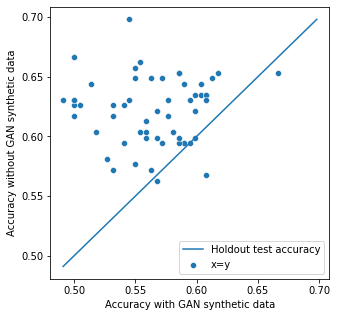}

}

\caption{\label{fig-gans-scatter-plot}Scatter plot comparing holdout
results from the best-performing models according to validation accuracy
with and without the inclusion of GAN-generated synthetic images in the
training data. A line marks where results would be equal.}

\end{figure}

While the generated images appear similar to the real CGR attractors,
they result in worse performance. We took the parameter sets that
yielded the highest validation accuracy on the real ArabidopsisPPI
dataset and retrained them with 2,000 true and 2,000 false synthetic
samples included. Figure~\ref{fig-gans-scatter-plot} plots the holdout
accuracy, and shows that only three of the models performed marginally
better with the synthetic data included.

\hypertarget{cgrs-contain-k-mer-count-features-that-indicate-taxonomic-ranks-introducing-potential-bias}{%
\subsection{\texorpdfstring{CGRs contain \(k\)-mer count features that
indicate taxonomic ranks, introducing potential
bias}{CGRs contain k-mer count features that indicate taxonomic ranks, introducing potential bias}}\label{cgrs-contain-k-mer-count-features-that-indicate-taxonomic-ranks-introducing-potential-bias}}

We continued the experiments into our approach to CGR-based PPI
prediction by attempting to repeat the GRAND and architecture search
processes using data taken from the HPIDB database(Ammari et al. 2016).
At 3,282 non-redundant pairs, this set is larger than ArabidopsisPPI
(1,106 pairs) and EffectorPPI (183 pairs). It also includes more variety
among hosts; while EffectorPPI is limited to interactions between
Arabidopsis thaliana and pathogens, HPIDB includes 66 host species.

Initial results were promising, with validation accuracy reaching
78.45\%. However, we questioned whether we could be confident that the
model was learning patterns relating to the actual interactions, or
whether it was biased by species information. The positive samples
experimentally validated interactions between hosts and pathogens from
various species, whereas the negative data is randomly drawn from all
hosts and all species. As our chosen CGR format represents the coding
sequence that codes for a protein rather than the amino acid sequence,
our data includes features that have been used in the field of
metagenomics to characterise and distinguish genomes such as GC-content,
codon usage and di-, tri- and tetranucleotide frequencies. While it is
possible that the model may learn subtleties of interaction that
preclude, for example, an Arabidopsis R-protein interacting with an
influenza effector, it may also have simply learned to reject the
interaction because the host does not have the sequence characteristics
of an animal.

To explore this hypothesis we masked our dataset, replacing each CGR
with a one-hot encoding of the species from which the gene originated.
We trained and tested a k-nearest neighbour classifier on the masked
data and from the species alone, achieved a holdout test accuracy of
87.67\%. This indicates that bias is present in the species
distributions of the data, from which even a simple model can appear to
learn to accurately predict interactions.

\begin{figure}

\begin{minipage}[t]{0.33\linewidth}

{\centering 

\raisebox{-\height}{

\includegraphics{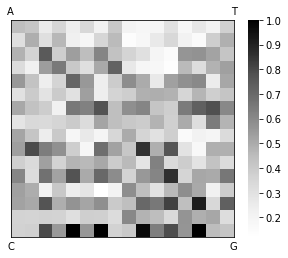}

}

}

\subcaption{\label{fig-rice-cgr}\emph{Oryza sativa}}
\end{minipage}%
\begin{minipage}[t]{0.33\linewidth}

{\centering 

\raisebox{-\height}{

\includegraphics{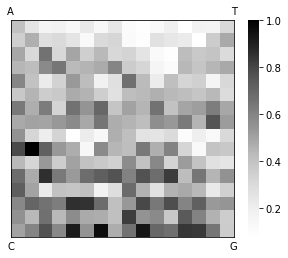}

}

}

\subcaption{\label{fig-magnaporthe-cgr}\emph{Magnaporthe oryzae}}
\end{minipage}%
\begin{minipage}[t]{0.33\linewidth}

{\centering 

\raisebox{-\height}{

\includegraphics{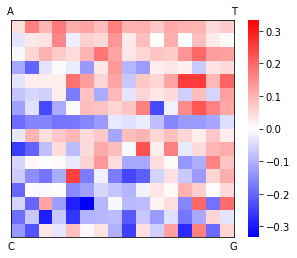}

}

}

\subcaption{\label{fig-magnaporthe-rice-diff}Difference}
\end{minipage}%

\caption{\label{fig-rice-magnaporthe-group}CGRs showing the entire
genome for (a) a host (\emph{Oryza sativa}) and (b) a pathogen
(\emph{Magnaporthe oryzae}) species, scaled to values between 0 and 1.
(c) shows the difference between the two plots, where positive (red)
values indicate that the 4-mer was more represented in the \emph{Oryza
sativa} genome and negative (blue) values indicate that the 4-mer was
more represented in the \emph{Magnaporthe oryzae} genome.}

\end{figure}

In order to learn from this bias, our models would have to be able to
know from which species a gene represented in a CGR had originated. We
therefore wanted to see if CGR attractors from different species'
genomes were noticeably different. We compared the sequences of two
different genomes, building CGR attractors for all CDS sequences in the
\emph{Magnaporthe oryzae} and \emph{Oryza sativa} genomes as shown in
Figure~\ref{fig-rice-magnaporthe-group}. We compiled all CGR attractors
into a single image for each of the Magnaporthe
(Figure~\ref{fig-magnaporthe-cgr}) and rice (Figure~\ref{fig-rice-cgr})
genomes. There are some visible differences for individual cells such as
strong representation of individual 4-mers such as ``CTCC'' in Oryza and
``CGAG'' in Magnaporthe. There also appears to be a more general
pattern, in that the Oryza CGR is strongest at certain positions along
the A-G diagonal and C-G horizontal lines, and Magnaporthe is generally
darker at most other points below the A-G diagonal.
Figure~\ref{fig-magnaporthe-rice-diff} shows this pattern via a heatmap
of the differences between the two plots where the scaled pathogen
combined CGR was subtracted from the scaled host combined CGR.

\hypertarget{cgr-cnns-can-learn-to-perform-taxonomic-classification}{%
\subsection{CGR CNNs can learn to perform taxonomic
classification}\label{cgr-cnns-can-learn-to-perform-taxonomic-classification}}

The HPIDB dataset bias and the differing features between genomes cast
doubt on our intended goal of predicting the interactions in the HPIDB
dataset. However, the ability to distinguish between taxonomic classes
also has potential applications. We wanted to investigate whether CNNs
trained with CGR attractors would work well at taxonomic classification
tasks as, when built and trained, our resulting models could provide a
relatively light-weight alternative to existing taxonomic classification
models. We performed two experiments to this end. In the first, we
classified gene CGRs according to their superkingdom to see how well the
model could give a broad overview of an unknown gene's origin. In the
second, we classified read CGRs according to whether they came from a
specific host or pathogen organism, as this may be helpful as an early
step in assembling sequences taken from sequencing an infected plant.

\begin{figure}

{\centering \includegraphics{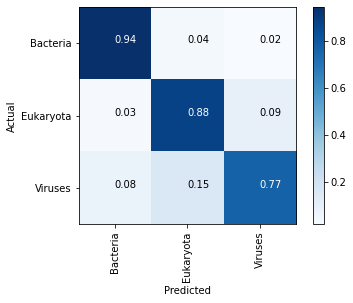}

}

\caption{\label{fig-species-prediction-conf-matrix}Confusion matrix from
a CNN model which uses CGR inputs to predict super kingdom from CDS
region. Data was extracted from the HPIDB dataset and consists of 757
bacteria samples, 1,745 eukaryote samples and 124 virus samples}

\end{figure}

The best-performing model architectures found during the PPI prediction
model search were not directly applicable to the classification problems
due to differences in input and output of the tasks. We performed new
parameter searches for each taxonomic classification task, where each
model would take in one CGR as input, and output predictions of class
membership.

Figure~\ref{fig-species-prediction-conf-matrix} shows the holdout test
performance of a CNN trained on CGRs taken from our HPIDB dataset and
labelled according to superkingdom. The mean class accuracy was 86.17\%.
The 757 bacteria samples in the test data were generally predicted well,
with 94\% correctly identified. There was some confusion between
eukaryota and viruses; the model achieved class accuracies of 88\% for
the more abundant eukaryotic test samples and 77\% for the viruses.

\begin{figure}

{\centering \includegraphics{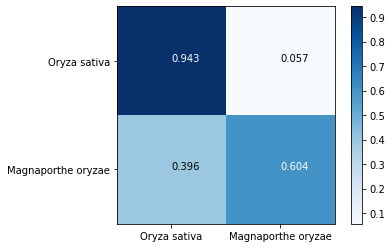}

}

\caption{\label{fig-magnaporthe-rice-conf-matrix}Confusion matrix
showing the results of a CNN model attempting to classify synthetic
reads}

\end{figure}

Read classifier results are shown in
Figure~\ref{fig-magnaporthe-rice-conf-matrix}. While the \emph{Oryza
sativa} reads were classified correctly in 94.3\% of cases, the
prediction of \emph{Magnaporthe oryzae} reads were only correct in
60.4\% of reads tested, giving a mean class accuracy of 77.37\%. While
the model did learn to discriminate between the two classes to a degree,
the use of CGRs and a CNN did not seem to achieve more than is possible
without them. We trained a linear logistic regression model on the same
data, extracting the \(k\)-mer counts from the CGR attractors into
feature vectors. It performed similarly with a mean class accuracy of
76.90\%, where the model was better at classifying the Oryza reads.

\hypertarget{discussion}{%
\section{Discussion}\label{discussion}}

The task of PPI prediction is challenging, and machine learning
approaches require careful consideration of the data that is used.
Machine learning requires abundant data but including all known
interactions introduces representation bias. However, the interactions
in PPI datasets often consist of many interactions for the same
proteins, so eliminating redundancy significantly reduces the amount of
available data. We have developed an algorithm for eliminating redundant
data while maintaining more samples than random discarding.

From a comprehensive architecture search we have developed a model that
is able to learn PPI prediction from small, non-redundant datasets. When
applied to the benchmark PPIPre datasets, we obtained results higher
than previously achieved for both the easier Negatome version and the
difficult RecombinePairs version of the data.

Smaller values of \(k\) result in better prediction accuracy despite the
coarser resolution information they contain. It may be that shorter
sequences are sufficient to discriminate between classes making longer
sequences redundant. It may also be that as the CGRs double in height
and width with each increment of \(k\), larger attractors contain too
many more features than there are samples in the dataset, resulting in
more overfitting. Finally, higher values of \(k\) mean that there are
more possible \(k\)-mers, and so fewer cells in a CGR will contain data.
CNN models are known to perform badly on sparse data.

The arrangement of \(k\)-mer counts within CGRs has assisted with this
performance, as subsets of the search area that do not include
convolutional layers do not obtain the same maximum accuracy.
Furthermore, machine learning methods without neural networks that use
vectors of \(k\)-mer counts as their input without regard to order do
not perform as well as the CGR CNN models. Whether there are different
arrangements within a CGR-like structure that would lead to higher
performance remains an open question.

We explored a method of augmenting a CGR dataset with synthetic
sequences by synonymous substitution, theorising that as the different
CDS sequences would encode the same protein sequence, homology and
likely identical behaviour would follow. While augmenting images to
increase the variety of input often increases the accuracy of computer
vision models, our experiments showed that this means of augmentation
generally does not improve our model at its PPI prediction task. The
synonymous substitutions appear to offer some improvement when the model
performed particularly badly on the validation data without augmented
training data. We consider that this may be the result of a slight
reduction in overfitting that caused the original poor performance.
However, the improvements gained by adding the synthetic images were
much weaker than improvements gained by choosing more suitable
architecture, and when performance was high without synonymous
substitutions, their addition made validation accuracy worse. We also
note that depending on the intended task of the model, the substitutions
may lose codon usage or tetranucleotide frequency biases that may have
been useful features.

GANs do not provide even this small potential for improvement, but
almost always worsened performance. The GAN model itself may require
more data in order to learn enough to produce synthetic images with
sufficient accuracy and variety. It may also be the case that while the
generated CGR attractors seem to share qualities with the real images to
the human eye, they lack important distiguishing features needed for
class distinction.

The question of suitable augmentations for deep learning of genetic data
therefore remains open.

We have also shown that machine learning methods using \(k\)-mer counts
to predict protein-protein interactions could introduce bias if positive
and negative samples use different species pairings. \(k\)-mer counts
can be sufficient to identify taxonomic classifications, and uneven
species pairings between positive and negative data can be sufficient to
separate the data. Our models for high-level taxonomic classification
were able to learn features to discriminate between genomes with
accuracy that is high, albeit not comparable with existing weightier
methods. In addition, linear logistic regression using \(k\)-mer counts
achieved similar performance, indicating that for these tasks the CGR
arrangement and CNN architecture do not contribute to learning. Dick and
Green (2020)'s comprehensive study uses several approaches to building
CGRs from proteins, building models to identify the source organisms of
protein source identification. While their predictions were better than
random chance, they described these results as ``modest''. Our taxonomic
experiments, which look at broader classes, are not directly comparable
but generally perform well and achieve \textgreater90\% accuracy in the
prediction of some classes. Our CGRs, and the \(k\)-mers we use for
logistic regression, are built with the benefit of the true knowledge of
the genetic coding sequence, which may underline the importance of
features such as nucleotide pattern frequency and codon usage in
taxonomic classification tasks, and the relative difficulty of identify
source organism from protein sequence and from genetic sequence.

To COVID. Which made this all a lot more fiddly than it needed be.

We would like to thank George Deeks and the Norwich Bioscience
Institutes' Research Computing department for their assistance with the
High Performance Computing cluster. We would also like to thank
Volodymyr Chapman for compiling the ArabidopsisPPI data, Anthony Duncan
for helpful discussions about metagenomics and Bethany Nichols for
naming GRAND. RV was supported by BBSRC grant number BB/V509267/1. This
work was supported by Wave 1 of The UKRI Strategic Priorities Fund under
the EPSRC Grant EP/T001569/1, particularly the ``AI for Science'' theme
within that grant \& The Alan Turing Institute. DM was supported by The
Gatsby Charitable Foundation through a core grant to The Sainsbury
Laboratory.

\hypertarget{references}{%
\section*{References}\label{references}}
\addcontentsline{toc}{section}{References}

\hypertarget{refs}{}
\begin{CSLReferences}{1}{0}
\leavevmode\vadjust pre{\hypertarget{ref-ABBAS2021106279}{}}%
Abbas, Amreen, Sweta Jain, Mahesh Gour, and Swetha Vankudothu. 2021.
{``Tomato Plant Disease Detection Using Transfer Learning with c-GAN
Synthetic Images.''} \emph{Computers and Electronics in Agriculture}
187: 106279.
https://doi.org/\url{https://doi.org/10.1016/j.compag.2021.106279}.

\leavevmode\vadjust pre{\hypertarget{ref-Albert_2021_ICCV}{}}%
Albert, Paul, Mohamed Saadeldin, Badri Narayanan, Brian Mac Namee,
Deirdre Hennessy, Aisling O'Connor, Noel O'Connor, and Kevin McGuinness.
2021. {``Semi-Supervised Dry Herbage Mass Estimation Using Automatic
Data and Synthetic Images.''} In \emph{Proceedings of the IEEE/CVF
International Conference on Computer Vision (ICCV) Workshops}, 1284--93.

\leavevmode\vadjust pre{\hypertarget{ref-hpidb2016}{}}%
Ammari, Mais G., Cathy R. Gresham, Fiona M. McCarthy, and Bindu Nanduri.
2016. {``{HPIDB 2.0: a curated database for host--pathogen
interactions}.''} \emph{Database} 2016 (July).
\url{https://doi.org/10.1093/database/baw103}.

\leavevmode\vadjust pre{\hypertarget{ref-berardini2015arabidopsis}{}}%
Berardini, Tanya Z, Leonore Reiser, Donghui Li, Yarik Mezheritsky,
Robert Muller, Emily Strait, and Eva Huala. 2015. {``The Arabidopsis
Information Resource: Making and Mining the {`Gold Standard'} Annotated
Reference Plant Genome.''} \emph{Genesis} 53 (8): 474--85.

\leavevmode\vadjust pre{\hypertarget{ref-bromley1993signature}{}}%
Bromley, Jane, Isabelle Guyon, Yann LeCun, Eduard Säckinger, and Roopak
Shah. 1993. {``Signature Verification Using a" Siamese" Time Delay
Neural Network.''} \emph{Advances in Neural Information Processing
Systems} 6.

\leavevmode\vadjust pre{\hypertarget{ref-deng2009imagenet}{}}%
Deng, Jia, Wei Dong, Richard Socher, Li-Jia Li, Kai Li, and Li Fei-Fei.
2009. {``Imagenet: A Large-Scale Hierarchical Image Database.''} In
\emph{2009 IEEE Conference on Computer Vision and Pattern Recognition},
248--55. Ieee.

\leavevmode\vadjust pre{\hypertarget{ref-deschavanne2008exploring}{}}%
Deschavanne, P, and P Tuffery. 2008. {``Exploring an Alignment Free
Approach for Protein Classification and Structural Class Prediction.''}
\emph{Biochimie} 90 (4): 615--25.

\leavevmode\vadjust pre{\hypertarget{ref-dick2020chaos}{}}%
Dick, Kevin, and James R Green. 2020. {``Chaos Game Representations \&
Deep Learning for Proteome-Wide Protein Prediction.''} In \emph{2020
IEEE 20th International Conference on Bioinformatics and Bioengineering
(BIBE)}, 115--21. IEEE.

\leavevmode\vadjust pre{\hypertarget{ref-ding2019computational}{}}%
Ding, Ziyun, and Daisuke Kihara. 2019. {``Computational Identification
of Protein-Protein Interactions in Model Plant Proteomes.''}
\emph{Scientific Reports} 9 (1): 1--13.

\leavevmode\vadjust pre{\hypertarget{ref-dwibedi2017cut}{}}%
Dwibedi, Debidatta, Ishan Misra, and Martial Hebert. 2017. {``Cut, Paste
and Learn: Surprisingly Easy Synthesis for Instance Detection.''} In
\emph{Proceedings of the IEEE International Conference on Computer
Vision}, 1301--10.

\leavevmode\vadjust pre{\hypertarget{ref-eid2021systematic}{}}%
Eid, Fatma-Elzahraa, Haitham A Elmarakeby, Yujia Alina Chan, Nadine
Fornelos, Mahmoud ElHefnawi, Eliezer M Van Allen, Lenwood S Heath, and
Kasper Lage. 2021. {``Systematic Auditing Is Essential to Debiasing
Machine Learning in Biology.''} \emph{Communications Biology} 4 (1):
183.

\leavevmode\vadjust pre{\hypertarget{ref-unrealengine}{}}%
Epic Games. 2023. {``Unreal Engine.''}
\url{https://www.unrealengine.com}.

\leavevmode\vadjust pre{\hypertarget{ref-AlphaFold-Multimer2021}{}}%
Evans, Richard, Michael ONeill, Alexander Pritzel, Natasha Antropova,
Andrew Senior, Tim Green, Augustin Zidek, et al. 2021. {``Protein
Complex Prediction with AlphaFold-Multimer.''} \emph{bioRxiv}.
\url{https://doi.org/10.1101/2021.10.04.463034}.

\leavevmode\vadjust pre{\hypertarget{ref-Franceschetti:2017gc}{}}%
Franceschetti, Marina, Abbas Maqbool, Maximiliano J Jiménez-Dalmaroni,
Helen G Pennington, Sophien Kamoun, and Mark J Banfield. 2017.
{``{Effectors of Filamentous Plant Pathogens: Commonalities amid
Diversity.}''} \emph{Microbiology and Molecular Biology Reviews : MMBR}
81 (2): e00066--16. \url{https://doi.org/10.1128/MMBR.00066-16}.

\leavevmode\vadjust pre{\hypertarget{ref-gonzalez2020effectork}{}}%
González-Fuente, Manuel, Sébastien Carrère, Dario Monachello, Benjamin G
Marsella, Anne-Claire Cazalé, Claudine Zischek, Raka M Mitra, et al.
2020. {``EffectorK, a Comprehensive Resource to Mine for Ralstonia,
Xanthomonas, and Other Published Effector Interactors in the Arabidopsis
Proteome.''} \emph{Molecular Plant Pathology} 21 (10): 1257--70.

\leavevmode\vadjust pre{\hypertarget{ref-NIPS2014_5ca3e9b1}{}}%
Goodfellow, Ian, Jean Pouget-Abadie, Mehdi Mirza, Bing Xu, David
Warde-Farley, Sherjil Ozair, Aaron Courville, and Yoshua Bengio. 2014.
{``Generative Adversarial Nets.''} In \emph{Advances in Neural
Information Processing Systems}, edited by Z. Ghahramani, M. Welling, C.
Cortes, N. Lawrence, and K. Q. Weinberger. Vol. 27. Curran Associates,
Inc.
\url{https://proceedings.neurips.cc/paper/2014/file/5ca3e9b122f61f8f06494c97b1afccf3-Paper.pdf}.

\leavevmode\vadjust pre{\hypertarget{ref-haas2009genome}{}}%
Haas, Brian J, Sophien Kamoun, Michael C Zody, Rays HY Jiang, Robert E
Handsaker, Liliana M Cano, Manfred Grabherr, et al. 2009. {``{Genome
sequence and analysis of the Irish potato famine pathogen Phytophthora
infestans}.''} \emph{Nature} 461 (7262): 393--98.
\url{https://doi.org/doi:10.1038/nature08358}.

\leavevmode\vadjust pre{\hypertarget{ref-hamp2015more}{}}%
Hamp, Tobias, and Burkhard Rost. 2015. {``More Challenges for
Machine-Learning Protein Interactions.''} \emph{Bioinformatics} 31 (10):
1521--25.

\leavevmode\vadjust pre{\hypertarget{ref-hashemifar2018predicting}{}}%
Hashemifar, Somaye, Behnam Neyshabur, Aly A Khan, and Jinbo Xu. 2018.
{``Predicting Protein--Protein Interactions Through Sequence-Based Deep
Learning.''} \emph{Bioinformatics} 34 (17): i802--10.

\leavevmode\vadjust pre{\hypertarget{ref-ResNet}{}}%
He, Kaiming, Xiangyu Zhang, Shaoqing Ren, and Jian Sun. 2015. {``Deep
Residual Learning for Image Recognition.''} arXiv.
\url{https://doi.org/10.48550/ARXIV.1512.03385}.

\leavevmode\vadjust pre{\hypertarget{ref-btr708}{}}%
Huang, Weichun, Leping Li, Jason R. Myers, and Gabor T. Marth. 2011.
{``{ART: a next-generation sequencing read simulator}.''}
\emph{Bioinformatics} 28 (4): 593--94.
\url{https://doi.org/10.1093/bioinformatics/btr708}.

\leavevmode\vadjust pre{\hypertarget{ref-jeffrey1990chaos}{}}%
Jeffrey, H Joel. 1990. {``Chaos Game Representation of Gene
Structure.''} \emph{Nucleic Acids Research} 18 (8): 2163--70.
https://doi.org/\href{\%20https://10.1093/nar/18.8.2163}{
https://10.1093/nar/18.8.2163}.

\leavevmode\vadjust pre{\hypertarget{ref-jia2019ippi}{}}%
Jia, Jianhua, Xiaoyan Li, Wangren Qiu, Xuan Xiao, and Kuo-Chen Chou.
2019. {``iPPI-PseAAC (CGR): Identify Protein-Protein Interactions by
Incorporating Chaos Game Representation into PseAAC.''} \emph{Journal of
Theoretical Biology} 460: 195--203.

\leavevmode\vadjust pre{\hypertarget{ref-jones2019setting}{}}%
Jones, David T. 2019. {``Setting the Standards for Machine Learning in
Biology.''} \emph{Nature Reviews Molecular Cell Biology} 20 (11):
659--60.

\leavevmode\vadjust pre{\hypertarget{ref-jumper2021highly}{}}%
Jumper, John, Richard Evans, Alexander Pritzel, Tim Green, Michael
Figurnov, Olaf Ronneberger, Kathryn Tunyasuvunakool, et al. 2021.
{``Highly Accurate Protein Structure Prediction with AlphaFold.''}
\emph{Nature} 596 (7873): 583--89.

\leavevmode\vadjust pre{\hypertarget{ref-araport}{}}%
Krishnakumar, Vivek, Matthew R. Hanlon, Sergio Contrino, Erik S.
Ferlanti, Svetlana Karamycheva, Maria Kim, Benjamin D. Rosen, et al.
2014. {``{Araport: the Arabidopsis Information Portal}.''} \emph{Nucleic
Acids Research} 43 (D1): D1003--9.
\url{https://doi.org/10.1093/nar/gku1200}.

\leavevmode\vadjust pre{\hypertarget{ref-kristianingsih2021accurate}{}}%
Kristianingsih, Ruth, and Dan MacLean. 2021. {``Accurate Plant Pathogen
Effector Protein Classification Ab Initio with Deepredeff: An Ensemble
of Convolutional Neural Networks.''} \emph{BMC Bioinformatics} 22 (1):
1--22.

\leavevmode\vadjust pre{\hypertarget{ref-kumar2010hpidb}{}}%
Kumar, Ranjit, and Bindu Nanduri. 2010. {``HPIDB-a Unified Resource for
Host-Pathogen Interactions.''} In \emph{BMC Bioinformatics}, 11:1--6. 6.
BioMed Central.

\leavevmode\vadjust pre{\hypertarget{ref-le2014distributed}{}}%
Le, Quoc, and Tomas Mikolov. 2014. {``Distributed Representations of
Sentences and Documents.''} In \emph{International Conference on Machine
Learning}, 1188--96. PMLR.

\leavevmode\vadjust pre{\hypertarget{ref-li2006cd}{}}%
Li, Weizhong, and Adam Godzik. 2006. {``Cd-Hit: A Fast Program for
Clustering and Comparing Large Sets of Protein or Nucleotide
Sequences.''} \emph{Bioinformatics} 22 (13): 1658--59.

\leavevmode\vadjust pre{\hypertarget{ref-lochel2020deep}{}}%
Löchel, Hannah F, Dominic Eger, Theodor Sperlea, and Dominik Heider.
2020. {``Deep Learning on Chaos Game Representation for Proteins.''}
\emph{Bioinformatics} 36 (1): 272--79.

\leavevmode\vadjust pre{\hypertarget{ref-man2022review}{}}%
Man, Keith, and Javaan Chahl. 2022. {``A Review of Synthetic Image Data
and Its Use in Computer Vision.''} \emph{Journal of Imaging} 8 (11):
310.

\leavevmode\vadjust pre{\hypertarget{ref-mirza2014conditional}{}}%
Mirza, Mehdi, and Simon Osindero. 2014. {``Conditional Generative
Adversarial Nets.''} \emph{arXiv Preprint arXiv:1411.1784}.

\leavevmode\vadjust pre{\hypertarget{ref-Nalley:2016jp}{}}%
Nalley, Lawton, Francis Tsiboe, Alvaro Durand-Morat, Aaron Shew, and
Greg Thoma. 2016. {``{Economic and Environmental Impact of Rice Blast
Pathogen (Magnaporthe oryzae) Alleviation in the United States.}''}
\emph{PLoS ONE} 11 (12): e0167295.
\url{https://doi.org/10.1371/journal.pone.0167295}.

\leavevmode\vadjust pre{\hypertarget{ref-park2012flaws}{}}%
Park, Yungki, and Edward M Marcotte. 2012. {``Flaws in Evaluation
Schemes for Pair-Input Computational Predictions.''} \emph{Nature
Methods} 9 (12): 1134--36.

\leavevmode\vadjust pre{\hypertarget{ref-QIU20202428}{}}%
Qiu, Jiajun, Michael Bernhofer, Michael Heinzinger, Sofie Kemper, Tomas
Norambuena, Francisco Melo, and Burkhard Rost. 2020. {``ProNA2020
Predicts Protein--DNA, Protein--RNA, and Protein--Protein Binding
Proteins and Residues from Sequence.''} \emph{Journal of Molecular
Biology} 432 (7): 2428--43.
https://doi.org/\url{https://doi.org/10.1016/j.jmb.2020.02.026}.

\leavevmode\vadjust pre{\hypertarget{ref-pmlr-v139-rao21a}{}}%
Rao, Roshan M, Jason Liu, Robert Verkuil, Joshua Meier, John Canny,
Pieter Abbeel, Tom Sercu, and Alexander Rives. 2021. {``MSA
Transformer.''} In \emph{Proceedings of the 38th International
Conference on Machine Learning}, edited by Marina Meila and Tong Zhang,
139:8844--56. Proceedings of Machine Learning Research. PMLR.
\url{https://proceedings.mlr.press/v139/rao21a.html}.

\leavevmode\vadjust pre{\hypertarget{ref-rives2021biological}{}}%
Rives, Alexander, Joshua Meier, Tom Sercu, Siddharth Goyal, Zeming Lin,
Jason Liu, Demi Guo, et al. 2021. {``Biological Structure and Function
Emerge from Scaling Unsupervised Learning to 250 Million Protein
Sequences.''} \emph{Proceedings of the National Academy of Sciences} 118
(15): e2016239118.

\leavevmode\vadjust pre{\hypertarget{ref-grandtheftauto}{}}%
Rockstar Games. 2013. {``Grand Theft Auto v.''}

\leavevmode\vadjust pre{\hypertarget{ref-rost1999twilight}{}}%
Rost, Burkhard. 1999. {``Twilight Zone of Protein Sequence
Alignments.''} \emph{Protein Engineering} 12 (2): 85--94.

\leavevmode\vadjust pre{\hypertarget{ref-saleh2018effective}{}}%
Saleh, Fatemeh Sadat, Mohammad Sadegh Aliakbarian, Mathieu Salzmann,
Lars Petersson, and Jose M Alvarez. 2018. {``Effective Use of Synthetic
Data for Urban Scene Semantic Segmentation.''} In \emph{Proceedings of
the European Conference on Computer Vision (ECCV)}, 84--100.

\leavevmode\vadjust pre{\hypertarget{ref-shin2018medical}{}}%
Shin, Hoo-Chang, Neil A Tenenholtz, Jameson K Rogers, Christopher G
Schwarz, Matthew L Senjem, Jeffrey L Gunter, Katherine P Andriole, and
Mark Michalski. 2018. {``Medical Image Synthesis for Data Augmentation
and Anonymization Using Generative Adversarial Networks.''} In
\emph{Simulation and Synthesis in Medical Imaging: Third International
Workshop, SASHIMI 2018, Held in Conjunction with MICCAI 2018, Granada,
Spain, September 16, 2018, Proceedings 3}, 1--11. Springer.

\leavevmode\vadjust pre{\hypertarget{ref-VGG19}{}}%
Simonyan, Karen, and Andrew Zisserman. 2014. {``Very Deep Convolutional
Networks for Large-Scale Image Recognition.''} arXiv.
\url{https://doi.org/10.48550/ARXIV.1409.1556}.

\leavevmode\vadjust pre{\hypertarget{ref-singh2019rna}{}}%
Singh, Jaswinder, Jack Hanson, Kuldip Paliwal, and Yaoqi Zhou. 2019.
{``RNA Secondary Structure Prediction Using an Ensemble of
Two-Dimensional Deep Neural Networks and Transfer Learning.''}
\emph{Nature Communications} 10 (1): 5407.

\leavevmode\vadjust pre{\hypertarget{ref-Sperschneider:2015kb}{}}%
Sperschneider, Jana, Peter N Dodds, Donald M Gardiner, John M Manners,
Karam B Singh, and Jennifer M Taylor. 2015. {``{Advances and challenges
in computational prediction of effectors from plant pathogenic
fungi.}''} \emph{PLoS Pathogens} 11 (5): e1004806.
\url{https://doi.org/10.1371/journal.ppat.1004806}.

\leavevmode\vadjust pre{\hypertarget{ref-EfficientNet}{}}%
Tan, Mingxing, and Quoc V. Le. 2019. {``EfficientNet: Rethinking Model
Scaling for Convolutional Neural Networks.''}
\url{https://doi.org/10.48550/ARXIV.1905.11946}.

\leavevmode\vadjust pre{\hypertarget{ref-wang2019predicting}{}}%
Wang, Lei, Hai-Feng Wang, San-Rong Liu, Xin Yan, and Ke-Jian Song. 2019.
{``Predicting Protein-Protein Interactions from Matrix-Based Protein
Sequence Using Convolution Neural Network and Feature-Selective Rotation
Forest.''} \emph{Scientific Reports} 9 (1): 9848.

\leavevmode\vadjust pre{\hypertarget{ref-WARD2020103009}{}}%
Ward, Daniel, and Peyman Moghadam. 2020. {``Scalable Learning for
Bridging the Species Gap in Image-Based Plant Phenotyping.''}
\emph{Computer Vision and Image Understanding} 197-198: 103009.
https://doi.org/\url{https://doi.org/10.1016/j.cviu.2020.103009}.

\leavevmode\vadjust pre{\hypertarget{ref-wei2017improved}{}}%
Wei, Leyi, Pengwei Xing, Jiancang Zeng, JinXiu Chen, Ran Su, and Fei
Guo. 2017. {``Improved Prediction of Protein--Protein Interactions Using
Novel Negative Samples, Features, and an Ensemble Classifier.''}
\emph{Artificial Intelligence in Medicine} 83: 67--74.

\leavevmode\vadjust pre{\hypertarget{ref-xu2023comprehensive}{}}%
Xu, Mingle, Sook Yoon, Alvaro Fuentes, and Dong Sun Park. 2023. {``A
Comprehensive Survey of Image Augmentation Techniques for Deep
Learning.''} \emph{Pattern Recognition}, 109347.

\leavevmode\vadjust pre{\hypertarget{ref-yang2019critical}{}}%
Yang, Shiping, Hong Li, Huaqin He, Yuan Zhou, and Ziding Zhang. 2019.
{``Critical Assessment and Performance Improvement of Plant--Pathogen
Protein--Protein Interaction Prediction Methods.''} \emph{Briefings in
Bioinformatics} 20 (1): 274--87.

\leavevmode\vadjust pre{\hypertarget{ref-yang2020prediction}{}}%
Yang, Xiaodi, Shiping Yang, Qinmengge Li, Stefan Wuchty, and Ziding
Zhang. 2020. {``Prediction of Human-Virus Protein-Protein Interactions
Through a Sequence Embedding-Based Machine Learning Method.''}
\emph{Computational and Structural Biotechnology Journal} 18: 153--61.

\leavevmode\vadjust pre{\hypertarget{ref-btac749}{}}%
Yu, Dingquan, Grzegorz Chojnowski, Maria Rosenthal, and Jan Kosinski.
2022. {``{AlphaPulldown---a python package for protein--protein
interaction screens using AlphaFold-Multimer}.''} \emph{Bioinformatics}
39 (1). \url{https://doi.org/10.1093/bioinformatics/btac749}.

\leavevmode\vadjust pre{\hypertarget{ref-Yuliar:2015hp}{}}%
Yuliar, Yanetri Asi Nion, and Koki Toyota. 2015. {``{Recent trends in
control methods for bacterial wilt diseases caused by Ralstonia
solanacearum.}''} \emph{Microbes and Environments} 30 (1): 1--11.
\url{https://doi.org/10.1264/jsme2.ME14144}.

\end{CSLReferences}

\hypertarget{appendix}{%
\section{Appendix}\label{appendix}}

\hypertarget{research-methods}{%
\section{Research Methods}\label{research-methods}}

\hypertarget{dataset-preparation}{%
\subsection{Dataset preparation}\label{dataset-preparation}}

We have presented results from PPI prediction experiments using two
datasets, which we refer to as ArabidopsisPPI and EffectorPPI.

We built the larger dataset, ArabidopsisPPI, from 4,493 interactions
between between \emph{Arabidopsis thaliana} proteins. We restricted
these to the interactions that are supported by direct evidence from
TAIR(Berardini et al. 2015) and Araport(Krishnakumar et al. 2014).

The smaller dataset, EffectorPPI, focuses on interactions between
\emph{Arabidopsis thaliana} proteins and pathogen effectors. We
extracted this data from the EffectorK(González-Fuente et al. 2020)
collection of curated interactions.

We created a third dataset from interactions extracted from the
Host-Pathogen Interaction Database (HPIDB)(Kumar and Nanduri 2010),
which we intended to use for PPI prediction. After discovering that it
was potentially biased, we instead extracted the CGRs from it to
investigate taxonomic classification.

We made each of these collections of PPIs into CGR datasets using the
same process. After filtering for redundancy using GRAND, we generated
negative samples by pairing genes that were not selected for inclusion.
We drew pairs of genes at random from these unused sequences, and then
searched the original unfiltered data to ensure that there is no known
interaction between these two genes or any other genes with similar
sequences; sequences are considered similar if they have a nucleotide
sequence identity of 80\% or higher.

From the PPI data, we retrieve the CDS that encodes each protein and
built these into CGR attractors using \(k\)-mers with lengths 4 to 7. We
stack the attractors for each pair to create data with two channels.

\hypertarget{grand}{%
\subsection{GRAND}\label{grand}}

We applied GRAND to three datasets: ArabidopsisPPI, EffectorPPI, and the
host-pathogen interactions taken from the HPIDB. The process followed
was the same in each case.

Using the CD-Hit-Est program, we filtered CDS data to an 80\% nucleotide
sequence identity, which was the lowest threshold provided by the
software. This resulted in clusters of similar sequences. We built a
network graph, wherein nodes represented each CD-Hit cluster and edges
represented that there was at least one known PPI between proteins coded
for by sequences in the connected clusters. We then reduced the graph
until it was non-redundant. While nodes existed that had only one
neighbour, we would randomly select from them and find their one
neighbour, removing all of that neighbours' other edges. If there were
no such nodes but there were still nodes with degree higher than one, we
iterate across edges and, for each edge, take the sum of the degrees of
the connected nodes. Nodes that have degree 0 are removed. This process
continues until all remaining nodes have exactly degree one.

We generated negative samples by selecting pairs of remaining clusters
and, provided that there was no known PPI interaction between proteins
in the clusters, drew one sample at random from each cluster. The order
of the proteins in the ArabidopsisPPI pairs was not relevant, so
clusters were drawn at random. We drew host and pathogen clusters
separately when generating negative data for the the HPIDB and EffectorK
datasets to maintain a similar distribution of species within the
negative data as the positive data.

\hypertarget{model-architecture}{%
\subsection{Model architecture}\label{model-architecture}}

Architectures were implemented, trained and tested in Python 3.6, using
the Keras 2.2.4 and TensorFlow 1.11.0 libraries.

We explored parameter sets that varied details of the architecture and
of the optimization. For \(k\) values 4,5,6 and 7, we tested 1,000
parameter sets. Optimizer details consisted of the optimization
algorithm, learning rate and regularization parameters L1 and L2.
Architecture parameters controlled the presence or absense of three main
blocks and, where present, their number and shapes: the twinned ResNet
block(s) each input was passed through; a second round of ResNet blocks
after concatenation of the paired inputs; dense layers before the final
classification. Dropouts, filter widths and biases for each of the three
main blocks were also included.

\hypertarget{ensembling-method}{%
\subsection{Ensembling method}\label{ensembling-method}}

We began by compiling validation results from all models tested in the
ArabidopsisPPI parameter search. We took the top 25 parameter sets
according to validation accuracy and created each combination of up to
10 of these models. For each combination, we loaded each model's
checkpoint and found its classification of each sample in the validation
data. We assigned these validation samples a predicted class by taking
the individual models' predictions as votes.

\hypertarget{transfer-learning}{%
\subsection{Transfer learning}\label{transfer-learning}}

We applied four approaches to architecture selection and training for
the EffectorPPI 4-mer dataset. A search for a parameter set was
performed using EffectorPPI, finding the best-performing according to
validation accuracy. The remaining three methods all used the best ten
parameter sets according to the validation results on the larger
ArabidopsisPPI dataset, but trained in different ways: training on
EffectorPPI from a random initialisation; training initially on
ArabidopsisPPI and then training further on EffectorPPI; training
initially on ArabidopsisPPI, freezing all weights but those in the final
classification layer and then training this layer on EffectorPPI.

\hypertarget{synthetic-images}{%
\subsection{Synthetic images}\label{synthetic-images}}

\hypertarget{synonymous-substitutions}{%
\subsubsection{Synonymous
substitutions}\label{synonymous-substitutions}}

We used synonymous substitutions of nucleotides to generate new
sequences from existing pairs of interacting proteins. We would build a
new pair of sequences by iterating through both sequences, taking each
codon in turn and randomly selecting from all codons that code for the
same amino acid.

\hypertarget{generative-adversarial-nets}{%
\subsubsection{Generative Adversarial
Nets}\label{generative-adversarial-nets}}

\begin{figure}

\begin{minipage}[t]{0.50\linewidth}

{\centering 

\raisebox{-\height}{

\includegraphics[width=1.82292in,height=\textheight]{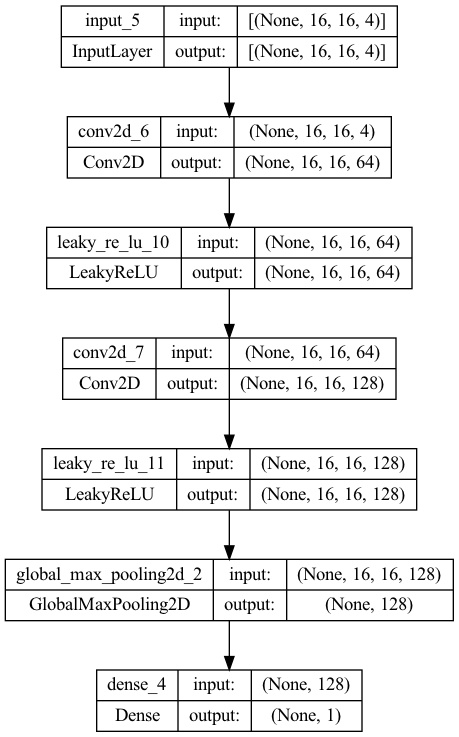}

}

}

\subcaption{\label{fig-gan-discriminator}Model architecture for GAN
discriminator sub-model}
\end{minipage}%
\begin{minipage}[t]{0.50\linewidth}

{\centering 

\raisebox{-\height}{

\includegraphics[width=1.82292in,height=\textheight]{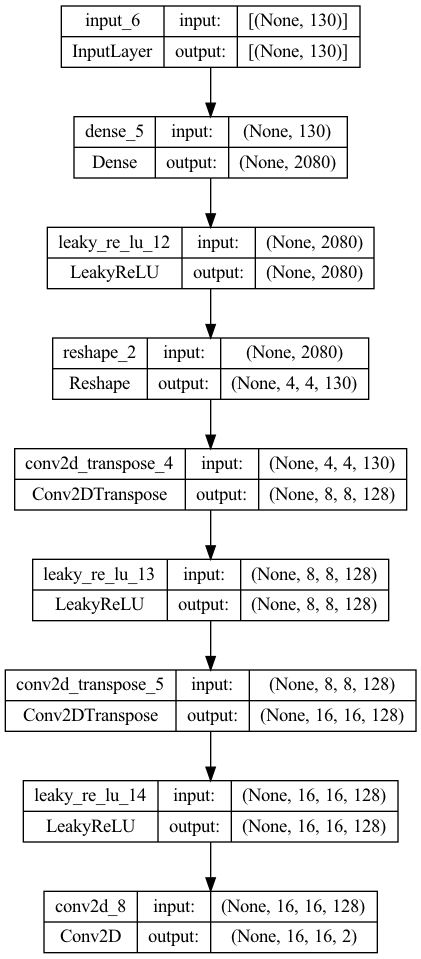}

}

}

\subcaption{\label{fig-gan-generator}Model architecture for GAN
generator sub-model}
\end{minipage}%

\caption{\label{fig-gan-architecture}GAN architecture}

\end{figure}

We implemented a simple cGAN architecture in TensorFlow and Keras,
following Keras' example of a cGAN model for generating MNIST digits.
The generator model takes the structure shown in
Figure~\ref{fig-gan-generator}, and the discriminator model architecture
is shown in Figure~\ref{fig-gan-discriminator}. Both sub-models were
trained with the Adam optimizer with beta parameters set to 0.5. The
learning rate for the generator was \(3\times10^{-4}\), for the
discriminator this was \(3\times10^{-3}\). Hyperparameters were chosen
by the researchers based on visual appearance; inappropriate
hyperparameters led to images that either were clearly lacking the
characteristics of CGRs or were almost identical within each class. All
hyperparameters were chosen before the synthetic data was tested on the
ArabidopsisPPI classification task. The GAN model was trained on
ArabidopsisPPI for 200 epochs, and then for short bursts of 10 epochs at
a time in between the generation of batches of fake images to further
encourage diversity among images of the same class.

\hypertarget{taxonomic-classification}{%
\subsection{Taxonomic classification}\label{taxonomic-classification}}

\hypertarget{classification-of-taxonomic-superkingdom}{%
\subsubsection{Classification of taxonomic
superkingdom}\label{classification-of-taxonomic-superkingdom}}

\begin{figure}

{\centering \includegraphics[width=1.82292in,height=\textheight]{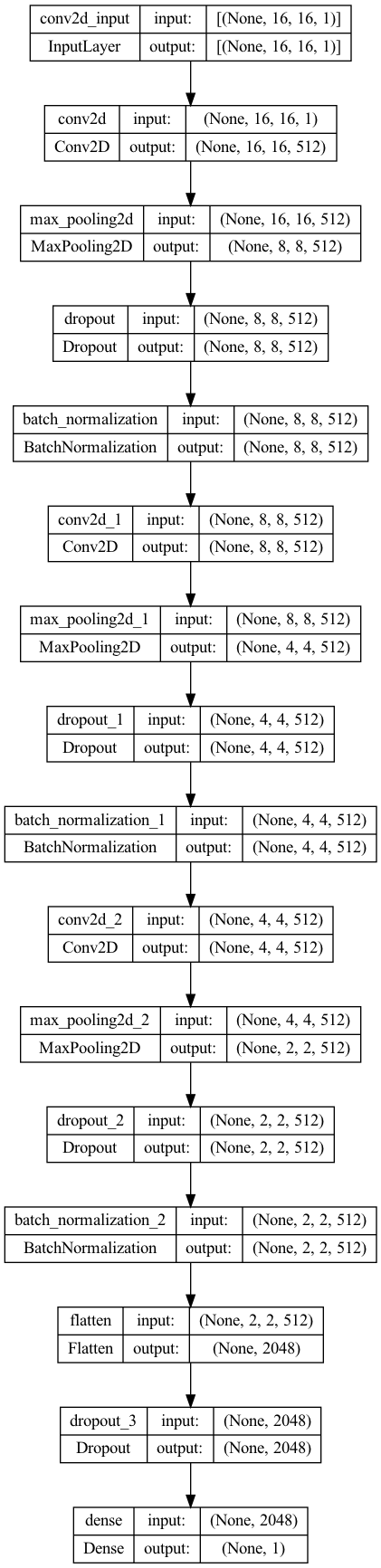}

}

\caption{\label{fig-superkingdom-model}Model architecture for
superkingdom taxonomic classifier}

\end{figure}

We took host-pathogen interactions from the HPIDB version 3.0
dataset(Kumar and Nanduri 2010) (Ammari et al. 2016) and reduced them to
3,282 non-redundant paired samples using GRAND. We paired each CGR with
a one-hot encoding of the superkingdom of the organism from which the
gene originated. As this data now has a single channel, we simplified
the model architecture search space and performed validation testing for
100 randomly generated parameter sets. Each parameter set was tested 5
times. The set with the highest mean validation categorical accuracy
used the architecture shown in Figure~\ref{fig-superkingdom-model}, was
optimized using Adam with a learning rate of 0.001, and was then
evaluated on the holdout data.

\hypertarget{hostpathogen-read-classification}{%
\subsubsection{Host/pathogen read
classification}\label{hostpathogen-read-classification}}

\begin{figure}

{\centering \includegraphics[width=1.82292in,height=\textheight]{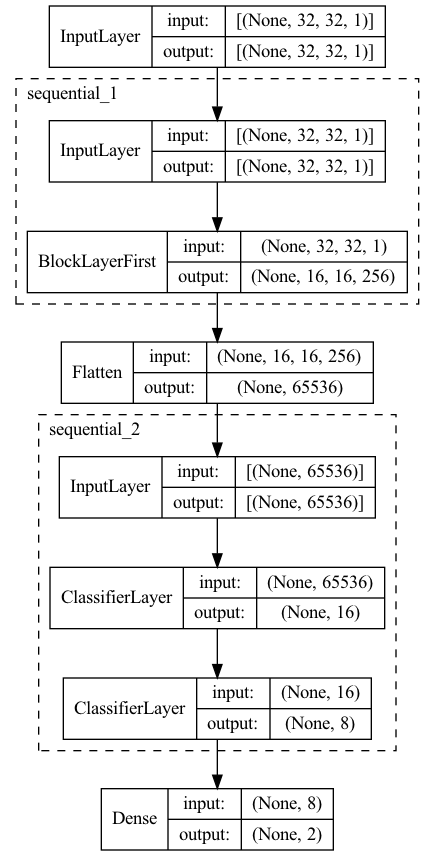}

}

\caption{\label{fig-hp-model}Model architecture for host/pathogen read
classifier}

\end{figure}

We took all 12,755 and 42,355 CDS sequences from the \emph{Magnaporthe
oryzae} and \emph{Oryza sativa} genomes respectively and created a
training dataset of CGR attractors. We also created a test dataset using
the ART software(Huang et al. 2011) to simulate Illumina single-end
sequencing reads, resulting in 107,272 Magnaporthe and 259,361 Oryza
reads. Variants of the CNN used for PPI prediction, using a single
ResNet segment instead of a twinned structure, were trained on the CDS
sequences and validated on synthetic reads represented as CGR attractors
with class weighting applied during training. The model with the highest
validation accuracy was then tested on the remaining synthetic reads'
CGRs. This model, which used 5-mers, is shown in
Figure~\ref{fig-hp-model}. The hyperparameter search also included
optimization parameters; the selected parameter set used the SGD
optimizer with a learning rate of \(1.5\times10^{-3}\). L1 and L2
regularisation were applied at rates \(1\times10^{-5}\) and
\({1\times10^{-3}}\).

\hypertarget{online-resources}{%
\section{Online Resources}\label{online-resources}}

Code is available from https://github.com/TeamMacLean/GRAND\_Code, and
data is hosted at https://zenodo.org/doi/10.5281/zenodo.10025321.

\end{document}